\documentclass[10pt,conference]{IEEEtran}
\IEEEoverridecommandlockouts

\usepackage{epsfig,rotating,setspace,latexsym,amsmath,epsf,amssymb,amsfonts,bm,theorem,cite,algorithm,graphicx,epsf,authblk,epstopdf,color,algpseudocode,bbm}

\usepackage{algorithm} 
\usepackage{algpseudocode} 
\usepackage{subfigure}
\usepackage{mathtools}
\usepackage{caption}

\allowdisplaybreaks

\normalsize

\ifCLASSINFOpdf
\else
\fi

\begin{document}

\title{Data Similarity-Based One-Shot Clustering for Multi-Task Hierarchical Federated Learning}
\author{Abdulmoneam Ali$~~~~~~\qquad$Ahmed Arafa\\Department of Electrical and Computer Engineering\\ University of North Carolina at Charlotte, NC 28223\\
$\quad$\emph{aali28@charlotte.edu}$\qquad$\emph{aarafa@charlotte.edu}
\thanks{This work was supported by the U.S. National Science Foundation under Grants CNS 21-14537 and ECCS 21-46099.}}

\markboth{IEEE Transactions on Communications}%
{Submitted paper}

\maketitle

\begin{abstract}

We address the problem of cluster identity estimation in a hierarchical federated learning setting in which users work toward learning different tasks. To overcome the challenge of task heterogeneity, users need to be grouped in a way such that users with the same task are in the same group, conducting training together, while sharing the weights of feature extraction layers with the other groups. Toward that end, we propose a \emph{one-shot} clustering algorithm that can effectively identify and group users based on their \emph{data similarity.} This enables more efficient collaboration and sharing of a common layer representation within the federated learning system. Our proposed algorithm not only enhances the clustering process, but also overcomes challenges related to privacy concerns, communication overhead, and the need for prior knowledge about learning models or loss function behaviors. We validate our proposed algorithm using various datasets such as CIFAR-10 and Fashion MNIST, and show that it outperforms the baseline in terms of accuracy and variance reduction.

\end{abstract}

\IEEEpeerreviewmaketitle

\section{Introduction}
Multi-Task Learning (MTL) and Personalized Federated Learning (PFL) have gained significant attention in recent years due to their potential in accelerating learning and improving generalization \cite{MTL-survey, marfoq2021federated}. In PFL, instead of learning a single global model, users seek to learn a personalized model while still leveraging the data and learning progress of other users, even if their specific model structures or parameters might differ. Likewise, MTL facilitates the simultaneous learning of multiple tasks using different models by leveraging the shared knowledge and information within the models' common representation. 

The inherent characteristics of multi-task learning (MTL) meet the requirements of Hierarchical Federated Learning (HFL). The core idea of HFL is to alleviate the communication bottleneck with the parameter server \cite{Hfl_kh, ali2023delay}. From a communication perspective, MTL offers the advantage of reducing the number of weights that need to be shared between users. By integrating MTL with HFL, groups of users can be associated with local parameter servers (LPSs), enabling them to jointly train their models within these LPSs. They can then communicate with the Global Parameter Server (GPS) to access the updated weights of common layers from other LPSs. This approach is expected to significantly alleviate the communication bottleneck encountered in Federated Learning (FL). The main challenge is how to efficiently design this multi-task hierarchical federated learning (MT-HFL) setting in a way that does not result in divergence or slow learning. 

\textbf{Related Work.} 
In \cite{clustered_FL}, the authors propose an alternating minimization algorithm that iterates between two phases: minimizing the loss function and identifying the cluster identities. They show that their algorithm converges, given a good initialization, with a strongly convex and smooth loss function. In \cite{NIPS2017}, the authors assume the existence of a matrix that captures the relationships between tasks. This matrix, either known beforehand or estimated during the training, is incorporated into the objective function along with a carefully chosen regularization term that depends on this task relation matrix. The training process then proceeds with this augmented objective function.  %
Reference \cite{Many-taskFL} evaluates the cosine similarity between the weights of each neural network layer among different pairs of users and considers this value as a metric to cluster users.   
The work in \cite{feature_hetero} focuses mainly on a multi-task linear regression model, and tackles feature heterogeneity using an augmented dataset-based approach. 
In \cite{RHFedMTL}, the authors assume a pre-clustered task structure, where each task is associated with a distinct cloud parameter server. Their objective is to determine the optimal number of local iterations per cloud server and global iterations, ensuring convergence within given resource budget and usage cost constraints.

\textbf{Contributions.}
Based on the literature, clustering approaches for MT-HFL can be broadly categorized into two types. The first type assumes that the model parameters for the same task are similar between users. It leverages this assumption by clustering users based on their model parameters, typically through iterative clustering algorithms that revisit the clustering task throughout the training process, as early-iteration weights may not be sufficiently informative for accurate cluster identification.
The second type involves designing a specific auxiliary loss function that addresses both feature heterogeneity and task-relation similarity.
Therefore the question we address in this work is as follows:
\begin{center}
\textit{Can we leverage feature heterogeneity among users to get a one-shot clustering algorithm for MT-HFL?}
\end{center}

The essence of this work is to turn the feature heterogeneity among users from a challenge into an {\it opportunity}. Our main objective is to efficiently cluster users among LPSs while preserving their data privacy and minimizing communication costs, and to do so {\it independently} of the model or the class of loss function at the users. Specifically, enabling each LPS to conduct training for a different task, determined by its associated users, and to collaborate with the other LPSs by sharing the common representation layers through the GPS.

To address the proposed question, we employ a variant of the data valuation method proposed in \cite{data_valuation}. Considering data valuation in clustering users has three advantages. Firstly, it provides one-shot clustering, with the decision being taken {\it before} training begins. Secondly, it provides an {\it optimum} user clustering solution that can be achieved in a distributed manner among users without intervention from the LPS or GPS. By optimum, we mean that our proposed algorithm successfully generates clusters that align with each user's specific task preferences, {\it as if it had prior knowledge of their individual tasks.} Thirdly, it is important to note that our approach does {\it not} rely on any prior knowledge regarding the model architecture, loss function, or the need for dataset augmentation. In summary, we mainly solve the concerns about one-shot clustering algorithms detailed in \cite{clustered_FL}. 
 
Our proposed algorithm can be summarized in two main steps: the first step estimates the feature similarity among users, as elaborated in Section \ref{sec:data-valuation}. Subsequently, we associate similar users with a specific LPS, as explained in Section \ref{sec:final_clustering_step}. Finally, we conduct a standard federated averaging training within each LPS, then share only subsets of the feature weights for aggregation with the GPS. To evaluate the effectiveness of our proposed clustering algorithm, we compare it with a random clustering approach that disregards the information regarding data similarity between users.

\textbf{Notation.} We use lowercase letters for scalars, bold lowercase letters for vectors, and bold uppercase letters for matrices throughout this paper.

\section{Data Similarity MT-HFL Clustering}\label{Sys_Model}

\subsection{System Model}
We consider an MT-HFL system with one GPS and a set of LPSs. The number of LPSs is equivalent to the number of tasks, $T$, that users in the system are aiming to learn. The set of users is defined as $\mathcal{N}=\{1,2,\dots,N\}$. Each user has its own data set $\{({\bm x}_l,y_l)\}_{l=1}^{n_
i}$, with ${\bm x}_l \in \mathbb{R}^{m} $ and $y_l \in \{1,\dots,C\}$, where $C$ is the number of classes and $n_i$ is the number of samples available at user $i$.

The main goal is to cluster users into $T$ clusters, without violating their data privacy, such that each cluster only contains users that seek to learn the same task. Users within the same cluster would then communicate with a dedicated LPS towards learning a model for their specific task. The LPSs, in turn, communicate with the GPS to learn a common layer representation to be shared among all clusters.

We assume that users who are looking to learn a specific task have most of their training samples drawn from that particular task in the training dataset. In addition, we assume that all users are honest and trustworthy.

\subsection{Data Similarity Estimation} \label{sec:data-valuation}

In order to cluster users into groups that have samples from the same task training dataset, we propose a variant of the data valuation approach in \cite{data_valuation} amenable to the FL setting. In particular, let us consider that user $i$ has its own raw data arranged in a matrix $\bm{X}_i \in \mathbb{R}^{n_i\times m}$. Our objective is to group user $i$ with user $j$ that has a data matrix $\bm{X}_j$ drawn from the same distribution as $\bm{X}_i$ (and therefore would be interested in learning the same task). The authors in \cite{data_valuation} propose a metric called data relevance that measures the similarity between the data of a seller and the data of a buyer, assisted by a broker. However, the assumptions made in their approach do not meet the FL data privacy requirements. We therefore propose a modified version of this metric by focusing on how to evaluate it in a distributed manner, without sharing users' data, without requiring any joint training among users, and with only at most \textit{one} iteration with the GPS.

Specifically, each user $i$ is required to compute an eigen decomposition of the Gram matrix of their data features. The Gram matrix is obtained by weighting the features with the number of samples of user $i$ as follows:
\begin{align}
\frac{1}{n_i}\Phi(\bm{X}_i)^{T}\Phi(\bm{X}_i),
\end{align}
where $\Phi(\cdot)$ is a feature mapping function belonging to $\mathbb{R}^{n_i\times d}$ with $d<m$.
The resulting eigen decomposition produces a set of eigenvalues arranged in a vector $\bm{\lambda}_i \in \mathbb{R}^d$ with their associated eigenvectors stacked in a matrix $\bm{V}_i \in \mathbb{R}^{d \times d}$.
Then, the users share their eigenvector matrix, $\bm{V}_i$, among themselves. Based on the matrix received, user $i$ can now estimate how much their data, $\bm{X}_i$, vary in the direction of each column of $\bm{V}_j$ from user $j$. This is achieved  by projecting the eigenvectors of other users onto their data and  evaluating the Euclidean norm as follows: 
\begin{align}
    \hat{\lambda}_{k}^{(j)}=\left\|\frac{1}{n_i}\Phi(X_i)^{T}\Phi(X_i) \bm{v}_{k}^{(j)}\right\|, ~\forall k \in [d], 
\end{align}
where $\bm{v}_{k}^{(j)}$ is the $k$th eigenvector of user $j$.

Reaching this step, one can expect that if two users, $i$ and $j$, have relatively similar features, i.e., their feature matrices lie close to each other in the feature space, then the values of their eigenvalues, both the original ($\bm{\lambda_i}$) and the projected ($\hat{\bm{\lambda}}_{j}$), should be relatively close as well. Therefore, each user $i$ can now compute the relevance \cite{data_valuation} of their data to user $j$ based on $\bm{\lambda}_i$ and $\bm{\hat{\lambda}}_j$ by evaluating:
\begin{align}
   \lambda_{k}^{(i,j)}=& \frac{\min\{\lambda_k^{(i)},\hat{\lambda}_k^{(j)}\}} {\max\{\lambda_k^{(i)},\hat{\lambda}_k^{(j)}\}}, ~\forall k \in [d], \label{eq:lambd(i,j)}\\
    r(i,j)=&\prod_{k=1}^{d} (\lambda_{k}^{(i,j)})^{\frac{1}{d}},\label{eq:r(i,j)}
\end{align}
where the $\max\{.,.\}$ in \eqref{eq:lambd(i,j)} is to normalize the value of $r(i,j)$.

After that, the GPS will receive from each user $i$ its data relevance estimates with respect to every other user $j \in \mathcal{N}$, as shown in \eqref{eq:r(i,j)}, and in turn estimates the average data similarity between each pair of users as follows:
\begin{align}\label{eq:avg_r}
    \bm{R}(i,j)=\frac{r(i,j)+r(j,i)}{2}, ~ \forall i, j \in \mathcal{N}.
\end{align}
Note that $\bm{R}$ is a symmetric $N\times N$ matrix.

\subsection{Clustering Decision} \label{sec:final_clustering_step}

Since $\bm{R}$ captures the pairwise data similarity/relevance between every user pair, the GPS adopts the Hierarchical Agglomerative Clustering algorithm (HAC) \cite{pml1Book} to cluster users among the LPSs. 

To better illustrate our idea, we show an example of matrix $\bm{R}$ in Table \ref{tab:rel_matrix} below. In this example, the CIFAR-10 dataset is distributed among five users such that users 1 and 2 have samples from class labels: $\{$plane, car, ship, truck$\}$, while users 3, 4, and 5 have samples from class labels: $\{$bird, cat, deer, dog, frog, horse$\}$.

\begin{table}[h]
\captionsetup{labelfont=normalfont,textfont=normalfont}
\centering
\begin{tabular}{c|ccccc}
\hline
Users & User 1 & User 2 & User 3 & User 4 & User 5 \\
\hline
User 1 & 1 & 0.97 & 0.31 & 0.31 & 0.32 \\
User 2 & 0.97 & 1 & 0.31 & 0.32 & 0.32 \\
User 3 & 0.31 & 0.31 & 1 & 0.97 & 0.98 \\
User 4 & 0.31 & 0.32 & 0.97 & 1 & 0.98 \\
User 5 & 0.32 & 0.32 & 0.98 & 0.98 & 1\\
\hline
\end{tabular}
\caption{Example of the data similarity matrix ${\bm R}$ on the CIFAR-10 dataset with two different tasks.}
\label{tab:rel_matrix}
\end{table}

In this example, the HAC algorithm initially assigns each user to a single cluster, resulting in five clusters. The algorithm proceeds to merge each close pair of users. For example, users 1 and 2 are very close, so they merge together. The same holds for users 4 and 5. Next, the algorithm merges user 3 into the cluster formed by users 4 and 5. The output of the algorithm is a binary tree known as a dendrogram. By varying the cutting height of this tree, different numbers of clusters can be obtained. The number of clusters needed in this example is 2, since there are two tasks to be learned. In general, the HAC algorithm is adjusted to produce $T$ clusters.

\begin{figure}[t]
    \centering
        \includegraphics[width=.95\linewidth]{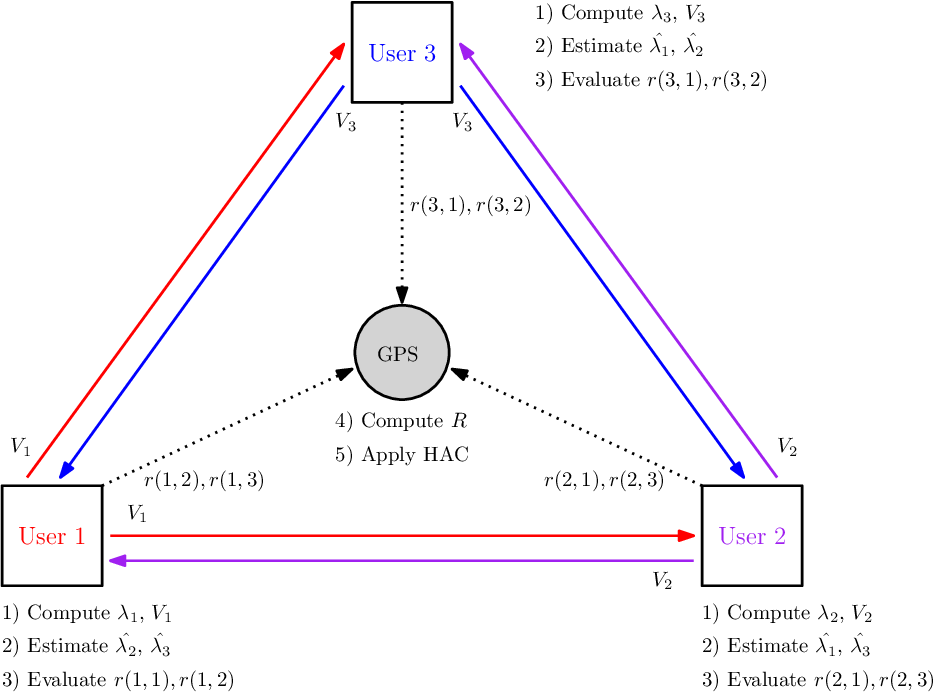}
    \caption{Illustration of our proposed algorithm.}
    \label{fig:clustering}
    \vspace{-.1in}
\end{figure}

\subsection{MT-HFL Training Procedure}

Once the assigned users are determined for each LPS, users can start the training process. Specifically, each LPS will individually execute a standard federated averaging algorithm (FedAvg) \cite{pmlr-v54-mcmahan17a} for a specified number of local iterations. Subsequently, they will share the weights of the common layers with the GPS. During each global communication round, the GPS aggregates the shared weights from the LPSs and broadcasts them back to initiate the next local iteration. These weighted average weights, combined with the LPSs' last weights for the remaining layers, serve as the starting point for the subsequent iteration. The training process continues until the predefined number of global iterations, $G$, is reached. The steps outlined above are summarized in Algorithms \ref{alg:main_algo} and \ref{alg:data_relevance} below.

Pictorially, our proposed approach is depicted in Fig.~\ref{fig:clustering}

\begin{algorithm}
    \caption{Main Algorithm}\label{alg:main_algo}

\begin{algorithmic}[1]
\State \textbf{Input}: number of tasks $T$; number of global iterations $G$
\State Execute Algorithm \ref{alg:data_relevance}: Data Similarity Clustering
\For {$r \in [G]$}
\For {$i \in [T]$}
\State Conduct FedAvg at each LPS
\State Share the weights of first common layers with GPS
\State GPS aggregates the shared weights and broadcasts them back to each LPS
\EndFor

\EndFor
\end{algorithmic}
 \end{algorithm}

\begin{algorithm}
    \caption{Data Similarity Clustering}\label{alg:data_relevance}

\begin{algorithmic}[1]

        \State Perform an eigenvalue decomposition:
		\For {$i \in \mathcal{N}$}
				\State  $\bm{\lambda}_i,\bm{V}_i=\text{eigen}(\frac{1}{n_i}\Phi(\bm{X}_i)^{T}\Phi(\bm{X}_i))$
                \State Share $\bm{V}_i$ with the other users 
        \EndFor
        \State Users compute the estimated the eigen values:  
        \For {$i \in \mathcal{N}$}
            \For {$j \in \mathcal{N}$}
				\State  $\hat{\lambda}_{k}=\|\frac{1}{n_i}\Phi(X_i)^{T}\Phi(X_i) \bm{v}_{k}^{(j)}\|$,  $~\forall k \in [d]$
                \State Compute $\lambda_{k} ^{(i,j)}=\frac{\min\{ \lambda_k^{(i)}, \hat{\lambda}_{k}\}}{\max \{\lambda_k^{(i)}, \hat{\lambda}_{k}\}}$, $~\forall k \in [d]$
                \State Apply \eqref{eq:r(i,j)} and share with the GPS
            \EndFor     
        \EndFor
		\State GPS uses \eqref{eq:avg_r} to compute the average similarity matrix $\bm{R}$:
         \For {$i \in \mathcal{N}$} 
               \State $ \bm{R}(i,j)=\frac{r(i,j)+r(j,i)}{2}$ and $\bm{R}(j,i) \gets  \bm{R}(i,j)$
         \EndFor
         
       \State Feed $\bm{R}$ to the HAC algorithm as input.
       \State Based on the desired number of clusters , $T$, and the output of the HAC algorithm, the memberships of the users to each cluster is obtained
	\end{algorithmic}
 \end{algorithm}

\begin{figure*}[t]
\centering
\subfigure[CIFAR-10 Task 1 LPS.]{\includegraphics[width=.425\linewidth]{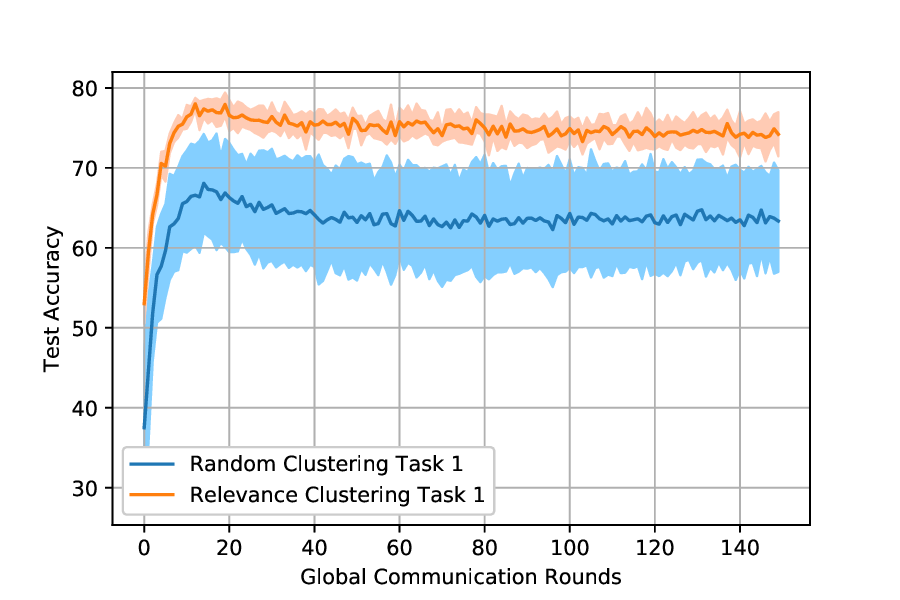}}
\subfigure[CIFAR-10 Task 2 LPS.]{\includegraphics[width=.425\linewidth]{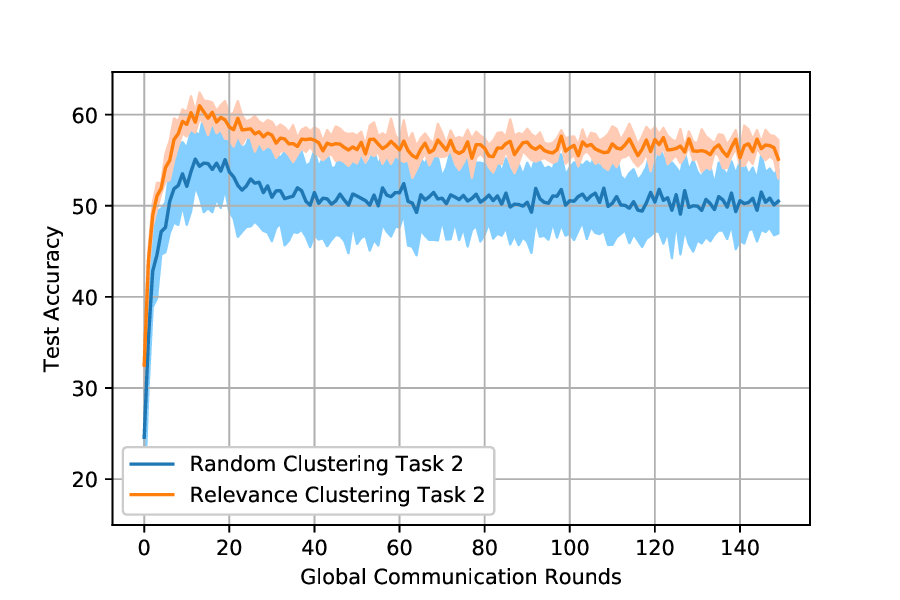}}
\caption{Performance on the CIFAR-10 dataset with two tasks/clusters.}
\label{fig:CIFAR10}
\vspace{-.2in}
\end{figure*}

\section{Experiments}\label{experiments}

We now present simulation results for the proposed data similarity-based clustering algorithm to demonstrate its superiority in achieving higher accuracy compared to the baseline approach. Furthermore, our approach exhibits a lower variance in the results. 
The subtle distinction between our proposed algorithm and the baseline is that the baseline ignores the similarity in the data between users and randomly clusters them.

\noindent \textbf{Datasets and Models.} We consider an image classification supervised learning task on the CIFAR-10 dataset \cite{krizhevsky2009learning} and the Fashion MNIST dataset \cite{fashion_mnist}. 
For CIFAR-10, a convolution neural network (CNN) is adopted with two 5x5 convolution layers, two 2x2 max pooling layers,  two fully connected layers with 120 and 84 units, respectively, ReLu activation, a final softmax output layer, and cross-entropy loss. In the case of Fashion MNIST, we adopt a multi-layer perceptron (MLP) architecture. The MLP comprises two fully connected layers: the first layer has dimensions (784, 32) and utilizes ReLU activation, followed by the second layer with dimensions (32, 10) and a log softmax activation. The loss function is the negative log likelihood loss. We conduct six experimental runs and calculate the average performance across these experiments. 

\noindent \textbf{Beating the Baseline.} In Fig. \ref{fig:CIFAR10}, we demonstrate the superiority of our proposed algorithm over random clustering using CIFAR-10.
We define two tasks: the first task involves learning  $\{$plane, car, ship, truck$\}$ labels, while the second task is to learn labels for  $\{$bird, cat, deer, dog, frog, horse$\}$. For each task, we distribute the corresponding labels among five users. Additionally, to introduce some level of label diversity and to test the robustness of our proposed algorithm, we assign 10\% labels from the other task to each user. The two LPSs are only sharing the weights of the two convolution layers with the GPS (common layers). 

\begin{figure*}[htbp]
\centering
\subfigure[FMNIST Task 1 LPS.]{\includegraphics[width=0.32\textwidth]{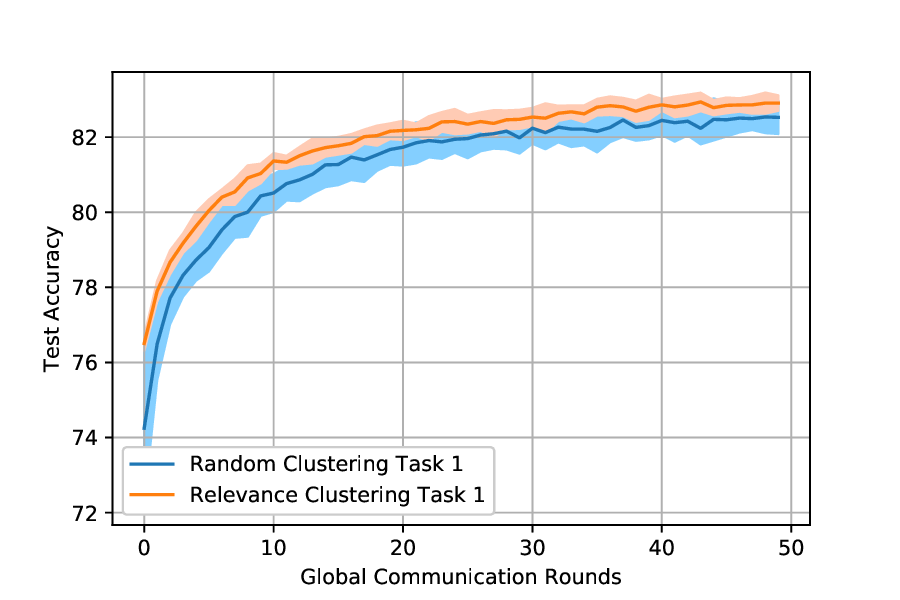}}
\subfigure[FMNIST Task 2 LPS.]{\includegraphics[width=0.32\textwidth]{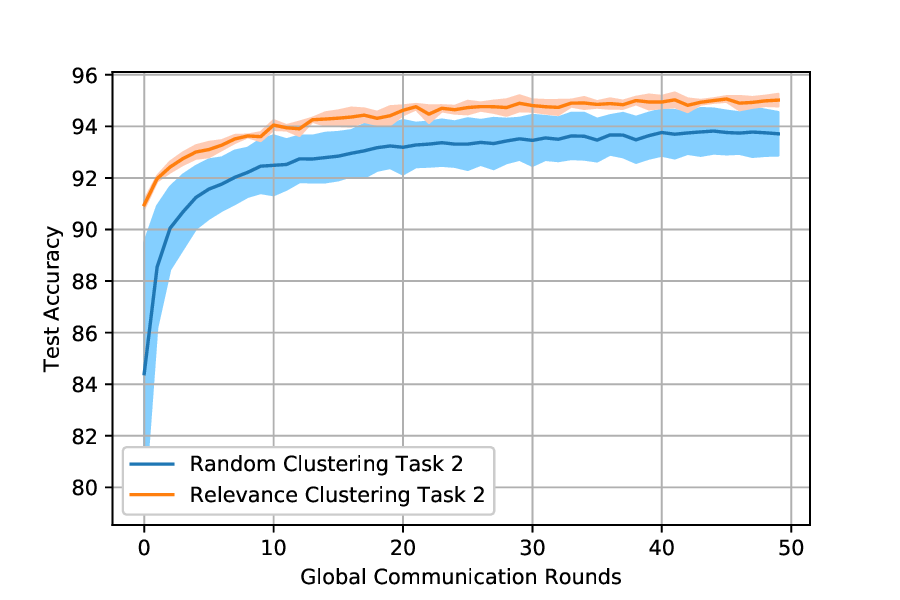}}
\subfigure[FMNIST Task 3 LPS.]{\includegraphics[width=0.32\textwidth]{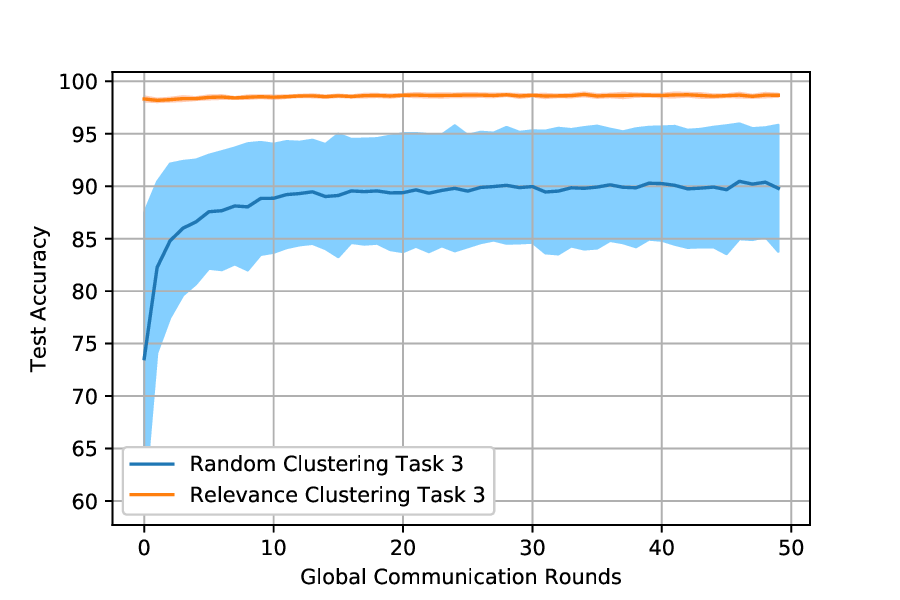}}

\caption{Performance on the Fashion MNIST with three tasks/clusters.}
\label{fig:FMNIST3Groups}
\vspace{-.1in}
\end{figure*}

To further validate the effectiveness of our proposed clustering algorithm, we conduct simulations involving three distinct tasks and an unbalanced distribution of task labels. In Fig. \ref{fig:FMNIST3Groups}, we show the superiority of the proposed algorithm using the Fashion MNIST dataset \cite{fashion_mnist}. We divide the dataset into three tasks: the first task focuses on learning clothes-related labels (Task 1), the second task is dedicated to learning shoe-related labels (Task 2), and the third task is to learn bag-related labels (Task 3). We have ten users: five of them have the majority of Task 1 labels, three users have the majority of Task 2 labels, and two users have Task 3 labels. Each user also has a minority of labels from other tasks.

In particular, the number of samples for each task is unbalanced. Specifically, Task 1 has a significantly larger number of samples compared to Task 2, and Task 3 has the smallest number of samples. In addition to that, only two users have the majority of Task 3 samples. This is why the performance of Task 3 is relatively lower in the random clustering baseline due to the lower possibility of choosing the two specific users related to this task out of the total of ten users. The significance of our proposed algorithm becomes evident in addressing such an unbalanced distribution of data. Unlike CIFAR-10, here we do not need a feature mapping $\Phi$, since the raw data itself is informative and has a lower dimension ($m=784$). For CIFAR-10, the raw data has a high dimension ($m=3072$), and are not inherently informative. Therefore, following the approach in the literature \cite{data_valuation}, \cite{feature_hetero}, \cite{Many-taskFL}, we utilize the convolution layers' weights of an ImageNet-pre-trained ResNet18 to evaluate the feature vector for each data point at users. {\it However, when users start training for learning, they start from random weights.}

\noindent \textbf{Dataset/Distribution Robustness.} Furthermore, our algorithm performs well when dealing with two distinct distributions. In particular, we investigated the similarity between users'labels using the CIFAR-10 and CIFAR-100 datasets in a three-user setting. User 1 has labels related to vehicles from CIFAR-10, while users 2 and 3 have CIFAR-100 data, with user 2 having labels representing vehicles, and user 3 having labels from the remaining classes of CIFAR-100.

The results, as shown in Table \ref{tab:cifar10-cifar100}, show the efficacy of our algorithm in successfully matching users with similar labels, even when such labels originate from different datasets. This showcases the algorithm's ability to handle diverse label distributions and facilitate effective collaboration among users.

\begin{table}[h]
\centering
\begin{tabular}{c|cc}
\hline
Users & User 2 & User 3 \\
\hline
User 1 & 0.62 & 0.39 \\
\hline
\end{tabular}
\caption{Similarity between users having different datasets: CIFAR-10 (User 1) and CIFAR-100 (Users 2 and 3).}
\label{tab:cifar10-cifar100}
\vspace{-.1in}
\end{table}

\begin{figure}[t]
    \centering
    \includegraphics[width=.8\linewidth]{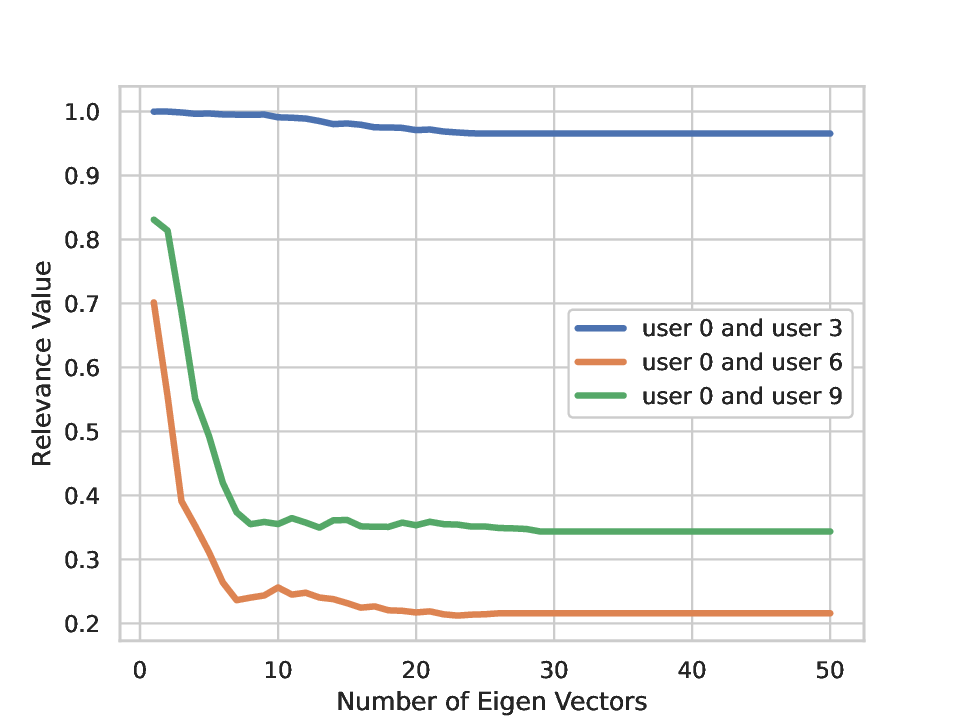}
    \caption{Impact of the number of shared eigenvectors on the relevance value.}
    \label{fig:n_eigens}
    \vspace{-.25in}
\end{figure}

\noindent \textbf{Communication Improvement.}
According to equation \eqref{eq:lambd(i,j)}, we observe that we need to discard the eigenvalues that are extremely small, as the multiplication will be highly drifted by such small values,  even if there is only one such eigenvalue. To illustrate this,  consider an example where $\hat{\lambda}_{100}=0.000001$ and $\lambda_{100}=1$. This observation, along with our goal to minimize the needed communication, motivates us to explore the optimal number of eigenvectors necessary to differentiate between users during the clustering step. Using the same setting as in the Fashion MNIST experiment, we select the first fifty eigenvectors corresponding to the fifty largest eigenvalues. As shown in Fig. \ref{fig:n_eigens}, it reveals that using only five eigenvectors, we can effectively cluster users with different tasks. This implies that instead of exchanging the entire matrix $\bm{V}$ with dimensions $(784 \times 784)$, users can exchange a much smaller matrix with dimensions $(5 \times 784)$ and still accurately compute their relatedness to each other. In Fig. \ref{fig:n_eigens}, Users 0 and 3 have samples from the same task,  which is different from the task associated with User 6 (having samples from Task 2) and user 9 (having samples from Task 3). This finding also emphasizes the power of our clustering algorithm over existing ones in the literature that mainly rely on the weights of the model, which are of much higher dimensions, at least on the order of thousands. 

\section{Conclusion}\label{conclusion}

A one-shot clustering data similarity-based algorithm is proposed in an MT-HFL setting, investigating the effects of clustering users based on their data on the overall accuracy of FL. Our proposed clustering algorithm enabled us to perfectly cluster users without violating their privacy. Furthermore, we achieved this with low-dimensional information exchange and independently of the learning model or the loss function class. For future investigations, we plan on studying the robustness of the proposed algorithm when the exchanged eigenvectors are noisy, in addition to exploring the possibility of adding an extra layer of privacy to our algorithm without compromising its clustering ability.

\ifCLASSOPTIONcaptionsoff
  \newpage
\fi

\bibliographystyle{unsrt}
\bibliography{Ali}

\end{document}